\documentclass[sigconf]{acmart}

\copyrightyear{2023}
\acmYear{2023}
\setcopyright{acmlicensed}\acmConference[MM '23]{Proceedings of the 31st
ACM International Conference on Multimedia}{October 29-November 3,
2023}{Ottawa, ON, Canada}
\acmBooktitle{Proceedings of the 31st ACM International Conference on
Multimedia (MM '23), October 29-November 3, 2023, Ottawa, ON, Canada}
\acmPrice{15.00}
\acmDOI{10.1145/3581783.3612045}
\acmISBN{979-8-4007-0108-5/23/10}

\AtBeginDocument{%
  \providecommand\BibTeX{{%
    \normalfont B\kern-0.5em{\scshape i\kern-0.25em b}\kern-0.8em\TeX}}}

\usepackage{color}
\usepackage{comment}

\usepackage{booktabs}

\usepackage[normalem]{ulem}
\usepackage{tabularx}
\usepackage[symbol]{footmisc}
\usepackage{multirow}
\usepackage{balance}

\begin{document}

\title{IDDR-NGP:~Incorporating Detectors for Distractor Removal with Instant Neural Radiance Field}



\author{Xianliang Huang}
\affiliation{%
  \institution{Fudan University}
  \country{Shanghai, China}
}
  \email{huangxl21@m.fudan.edu.cn}

\author{Jiajie Gou}
\affiliation{%
  \institution{Fudan University}
  \country{Shanghai, China}}
    \email{19307110343@fudan.edu.cn}
    
\author{Shuhang Chen}
\affiliation{%
  \institution{Fudan University}
  \country{Shanghai, China}}
  \email{22210240123@m.fudan.edu.cn}
  
\author{Zhizhou Zhong}
\affiliation{%
  \institution{Fudan University}
  \country{Shanghai, China}}
   \email{22210240402@m.fudan.edu.cn}

\author{Jihong Guan}

\affiliation{%
  \institution{Tongji University}
  \country{Shanghai, China}}
    \email{jhguan@tongji.edu.cn}
    
\author{Shuigeng Zhou}
\authornote{Corresponding author (\href{mailto:sgzhou@fudan.edu.cn}{sgzhou@fudan.edu.cn}).}
\affiliation{%
 \institution{Fudan University}
 \country{Shanghai, China}}
 \email{sgzhou@fudan.edu.cn}


\renewcommand{\shortauthors}{Xianliang Huang et al.}

\begin{abstract}
This paper presents the first unified distractor removal method, named IDDR-NGP, which directly operates on Instant-NPG. The method is able to remove a wide range of distractors in 3D scenes, such as snowflakes, confetti, defoliation and petals, whereas existing methods usually focus on a specific type of distractors.
By incorporating implicit 3D representations with 2D detectors, we demonstrate that it is possible to efficiently restore 3D scenes from multiple corrupted images.
We design the learned perceptual image patch similarity~( LPIPS) loss and the multi-view compensation loss (MVCL) to jointly optimize the rendering results of IDDR-NGP, which could aggregate information from multi-view corrupted images.
All of them can be trained in an end-to-end manner to synthesize high-quality 3D scenes. To support the research on distractors removal in implicit 3D representations, we build a new benchmark dataset that consists of both synthetic and real-world distractors. 
To validate the effectiveness and robustness of IDDR-NGP, we provide a wide range of distractors with corresponding annotated labels added to both realistic and synthetic scenes.
Extensive experimental results demonstrate the effectiveness and robustness of IDDR-NGP in removing multiple types of distractors. In addition, our approach achieves results comparable with the existing SOTA desnow methods and is capable of accurately removing both realistic and synthetic distractors.

\end{abstract}




\begin{teaserfigure}
  \includegraphics[width=\textwidth]{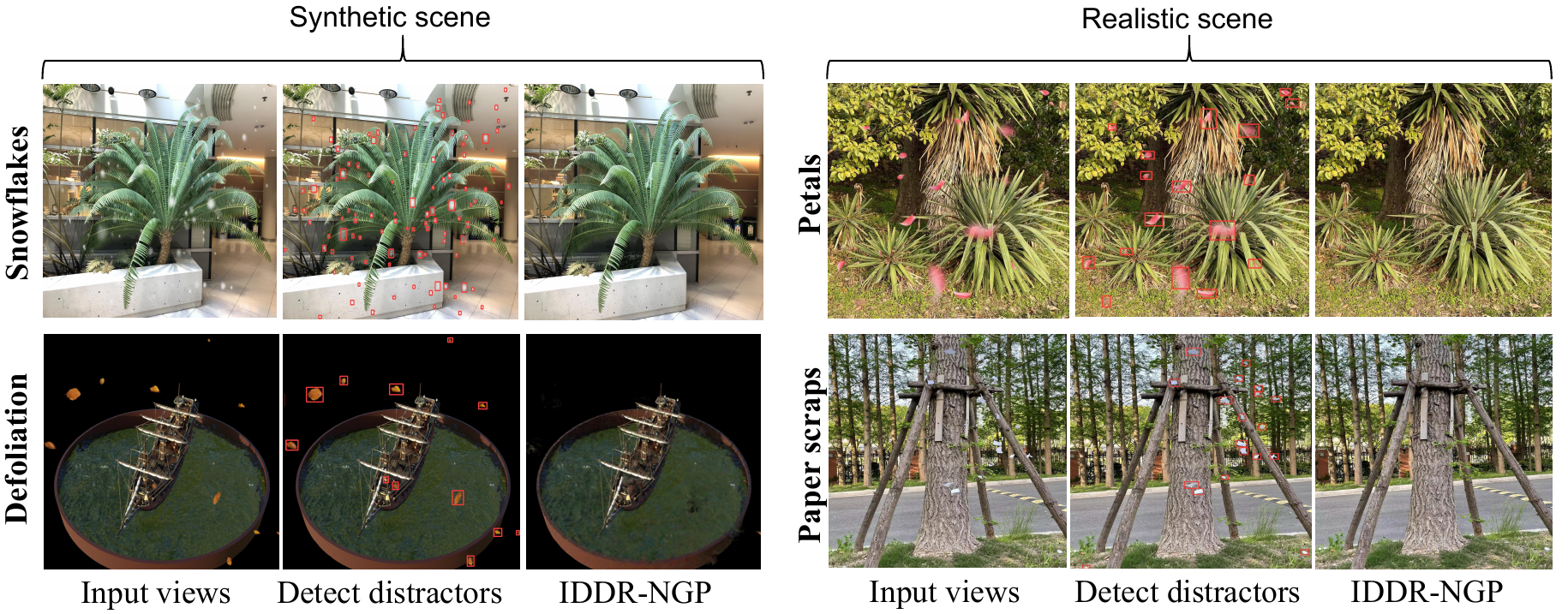}
  \caption{Experiments on both synthetic and realistic scenes with different types of distractors. Given the input views, we visualize the detected bounding boxes of distractors and the removal results of our method IDDR-NGP.}
  \label{fig:teaser}
\end{teaserfigure}

\maketitle
\section{Introduction}
With the growing popularity of computer vision applications, 
the removal of distractors in 3D scenes has become an essential step for achieving accurate and robust results of downstream CV tasks, and it is fundamental to many important applications such as robotics and autonomous driving. 
On the other hand, capturing both photos and videos in 3D scenes requires professionals to carefully plan the scenes and eliminate all unwanted objects. However, it is quite common to unintentionally capture visually distracting objects or elements that interfere with the analysis of the scenes, which we refer to as visual distractors. These distractors are regions in the images that divert attention away from the main objects, thereby reducing the overall image quality. 
As a result, the removal of 3D scene distractors can improve the accuracy and reliability of many CV tasks, including object recognition, tracking, and segmentation.


Despite its importance, removing various distractors in 3D scenes with a one-size-fits-all solution remains a challenging task. 
Specifically, the presence of multiple types of distractors requires handcrafted design features or well-designed models for removing each type of these distractors.
Compared to the removal of 2D distractors, removing objects in 3D scenes is more challenging due to the difficulty in data collection and annotation of 3D scenes.
Therefore, the field of 3D distractor removal poses a multitude of room for exploration.  
One effort towards achieving this goal is to automatically detect arbitrary types of distractors that need to be eliminated in the target scenes. On the other hand, removing these distractors demand a time-consuming editing process, which needs to manually select the target areas and obtain the corresponding mask information to complete the area. 

To address these challenges mentioned above, most existing works have tried various techniques, such as hand-crafted feature extraction and deep neural networks. These methods have achieved considerable success, however they have limited potential  to be generalized to a variety of complex 3D scene interferences. Furthermore, these methods often require specialized training for each type of interferences, making them inefficient and impractical. Some other existing methods require 3D point clouds as input. However, point cloud representations heavily rely on accurate data captured by 3D sensors.
Nevertheless, recent advances in Neural Radiance Fields (NeRF) \cite{mildenhall2021nerf} provide an effective alternative approach to extract highly semantic features of the underlying 3D scenes from 2D multi-view images.

In this paper, we propose a new method to incorporate 2D detectors for 3D scene distractor removal with instant Neural radiance field, which directly manipulates the implicit representation of a given 3D scene. Our method differs from previous works in that it is learned entirely from RGB images and camera poses, and is designed to be versatile and adaptable to a wide range of interference types. Furthermore, our method takes full advantage of the 3D information inherent in Instant-NGP and makes it a practical and efficient solution to real-world distractor removal. Specifically, given the output bounding boxes of detectors, the multi-resolution hash encodings are adapted to store the radiance field and the density. Then, we propose a multi-view compensation loss function and modify the rendering strategy of Instant-NGP to fully remove distractors inside bounding boxes.
The proposed method is called IDDR-NGP, the abbreviation of \textbf{I}ncorporating 2D \textbf{D}etectors for 3D scene \textbf{D}istractor \textbf{R}emoval with instant-\textbf{NGP}.  
We compare our method with different SOTAs on snow removal and create a dataset with different distractors added to a collection of multi-view images. 
In summary, as the first attempt to perform 3D distractor removal from multi-view images, this paper has the following contributions:
\begin{itemize}
\item We propose a novel distractor removal method called IDDR-NGP. To the best of our knowledge, this is the first method that exploits 2D detectors and Instant-NGP to efficiently remove unwanted distractors in 3D scenes.

\item  We modify the primitive Instant-NGP to avoid rendering distractors inside the bounding boxes, which enables IDDR-NGP to remove several types of distractors from arbitrary images. Furthermore, benefiting from the learned perceptual image patch similarity~( LPIPS) loss and the multi-view compensation loss (MVCL), we obtain more stable results in various scenes.

\item To verify the effectiveness of IDDR-NGP, we create a wide range of distractors with corresponding annotated labels, which are added to both realistic and synthetic scenes.

\item Extensive experiments demonstrate the effectiveness and superiority of IDDR-NGP in removing multiple types of distractors. Furthermore, our approach outperforms existing SOTA desnow methods and is capable of removing both realistic and synthetic distractors.

\end{itemize}

\begin{figure*}
\centering\includegraphics[width=1\textwidth]{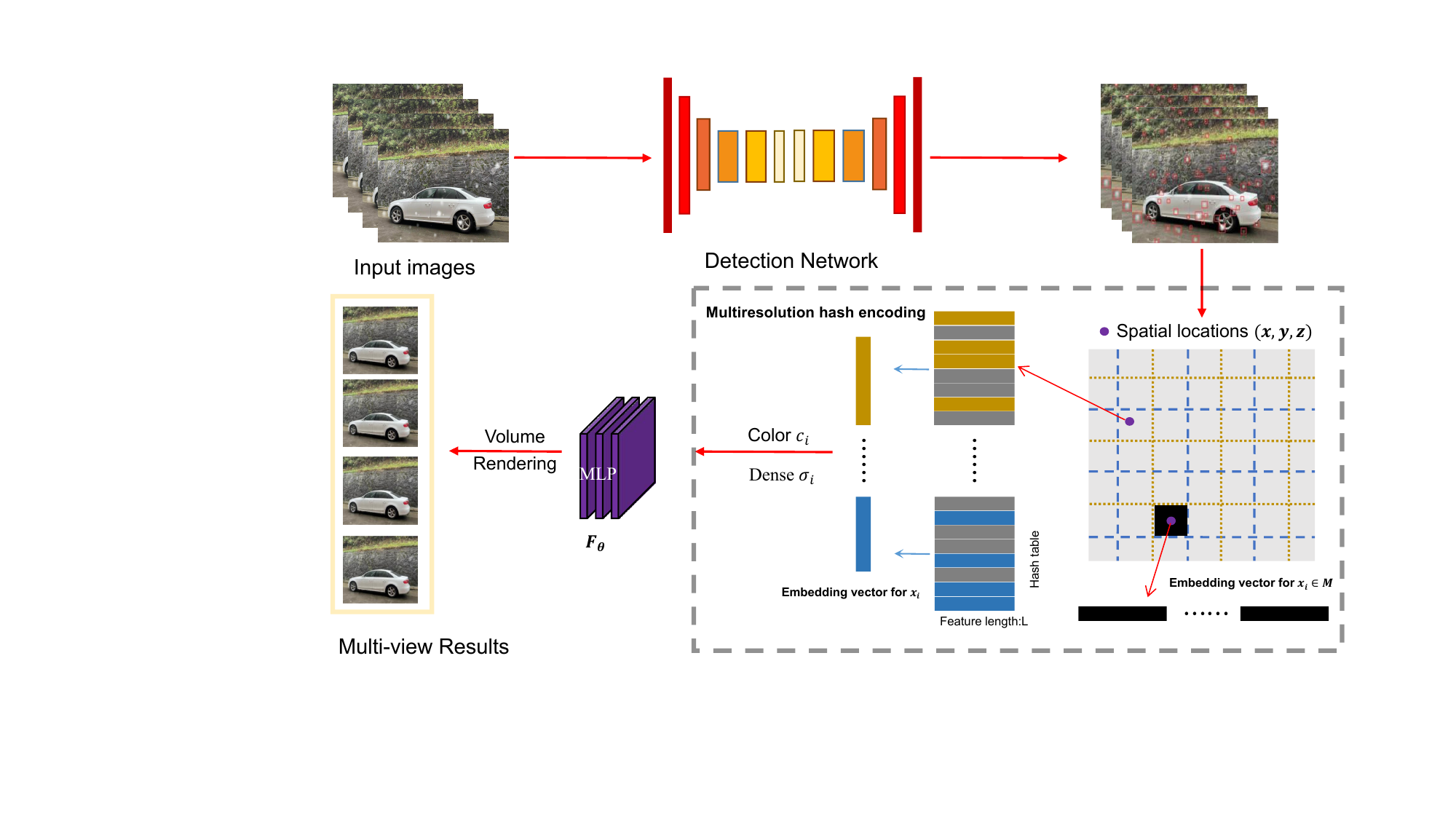}
\caption{The framework of our method IDDR-NGP. In the component of hash encoding, different colors denote different resolutions and the corresponding embedding vectors.}
\label{fig:frame}
\vspace{-0.4cm}
\end{figure*}

\section{Related Work}
\subsection{Neural Radiance Fields}
Recently, neural radiance fields (NeRF) have made great progress in computer vision in the last two years. Unlike the previous voxel, point cloud, and mesh-based methods, NeRF learns to represent the radiation fields in a 3D scene from a series of posed images by simply training a multi-layered perceptron~(MLP) network. The success of NeRF has inspired a variety of works for image processing~\cite{mildenhall2022nerf,huang2022hdr,pearl2022nan,ma2022deblur} or scene manipulation~\cite{niemeyer2021giraffe,liu2021editing,kania2022conerf,wang2022clip}.
However, these works do not address small and randomly distributed objects in realistic scenes. (e.g. leaves, petals, snowflakes and other obstacles). Training NeRF with contaminated images is an unexplored area, our work is the first that employs NeRF to restore both synthetic and realistic scenes with random distractors.

\subsection{Fast NeRF Training and Rendering}
Improving the training speed of NeRF has been received an extensive research in recent years. Several methods~\cite{sun2022direct,wang2022r2l} have been proposed
KiloNeRF~\cite{reiser2021kilonerf} leverages thousands of tiny MLPs rather than one large MLP to achieve real-time rendering and quicker training. Plenoxels~\cite{fridovich2022plenoxels} represent a scene as a sparse 3D grid with spherical harmonics, which enables optimization without neural components. 
Recently, Instant-NGP~\cite{muller2022instant} suggested storing voxel grid features in a multi-resolution hash table and using a spatial hash function to query these features, which considerably reduces the number of optimizable parameters. The position encoder of Instant-NGP adopts a neural network with trainable weight parameters and encoding parameters.
\subsection{Object Detection}
Currently, CNN-based object detectors can be divided into many types. One-stage detectors, e.g., YOLOX~\cite{ge2021yolox}, FCOS~\cite{tian2019fcos}, DETR~\cite{zhu2020deformable}, Scaled-YOLOv4~\cite{wang2021scaled}, EfficientDet~\cite{tan2020efficientdet}. Two-stage detectors, e.g., VFNet~\cite{zhang2021varifocalnet}, CenterNet2~\cite{zhou2021probabilistic}. Anchor-based detectors e.g., Scaled-YOLOv4~\cite{wang2021scaled}, YOLOv5~\cite{zhu2021tph}.
Anchor-free detectors, e.g., CenterNet~\cite{zhou2019objects}, YOLOX~\cite{ge2021yolox}.
Some detectors are specially designed for Drone-captured images like RRNet~\cite{chen2019rrnet}, CenterNet~\cite{zhou2019objects}. But from the perspective of components, they generally consist of two parts, an CNN-based backbone, used for image feature extraction, and the other part is detection head used to predict the class and bounding box for object. In addition, the object detectors developed in recent years often insert some layers between the backbone and the head, people usually call this part the neck of the detector. YOLOv5~\cite{zhu2021tph} is the most advanced improved model of the current YOLO series. It adopts a more efficient CBAM module and proposes Transformer Prediction Heads(TPH) for auxiliary head training, which has high accuracy and performance. Therefore, we choose YOLOv5 as the detection component of our framework.

\subsection{Removing Distractors from 3D Scenes}
A primary characteristic of distractors is that they attract our visual attention, so they are likely to be somewhat correlated with models of visual saliency. Another line of related work focuses on automatic image cropping [29, 21, 34]. While cropping can often remove some visual distractors, it might also remove important content. For instance, many methods just try to crop around the most salient object. Advanced cropping methods [34] also attempt to optimize the layout of the image, which might not be desired by the user and is not directly related to detecting distractors.  Removing distractors is also related to the visual aesthetics literature [16, 19, 23, 30] as distractors can clutter the composition of an image, or disrupt its lines of symmetry.
Image and video enhancement methods have been proposed to detect dirt spots, sparkles [17], line scratches [14]
and raindrops [8].
Another interesting work [28] focused on detecting and de-emphasizing distracting texture regions that might be more salient than the main object. All of the above methods are limited to a certain type of distractor or image content, but in this work we are interested in a more general-purpose solution.

\section{Methodology}
Given captured images $\mathcal{I} = \{I_1, I_2,...,I_k\}$ with multiple objects as distractors, we want to annotate all distractors, and then construct a model to generate scenes with these distractors. Firstly, we employ the YOlOv5~\cite{zhu2021tph} algorithm to identify all distractors $\mathcal{D}_k = \{D_1, D_2,...,D_i\}$ in image $I_k$ and obtain corresponding bounding boxes, which contain the coordinate of upmost and rightmost vertex ($v^{(i)}_x$,$v^{(i)}_y$) in the image plane. Secondly, we utilize simple scripts to set a default RGB value (96,96,96) to the identified YOLO coordinate regions, thus producing corresponding mask matrices.
Fig.~\ref{fig:frame} illustrates an overview of our pipeline. First, we provide some background in Sec \ref{sec:background}. Then, the basic principle of YOLOv5 algorithm is described in Sec \ref{sec:Yolov5}, which provides the high accuracy and real-time performance of detection results. Refined Multi-resolution hash encodings are introduced in Sec \ref{sec:hash}. Finally, we present the training strategy and loss function in Sec \ref{sec:loss}.
\subsection{Background}
\label{sec:background}
\noindent{\bf{Neural Radiance Field.}} Neural Radiance Field (NeRF)~\cite{mildenhall2021nerf} is a 5D MLP function $F_{\omega}$ that represents a static 3D scene. It calculates the radiance emitted in each direction $\mathbf{v}$ at each point $\mathbf{x}$ in space and a density at each point. NeRF uses a volume rendering strategy to render images based on the neural radiance field.
During rendering, NeRF samples a ray $\boldsymbol{\gamma} = \mathbf{o} + t\mathbf{v}$ per pixel and then calculates the color of the pixel using the following volume rendering strategy~\cite{lombardi2019neural}:
\begin{equation}
\centering
\mathbf{C}(\boldsymbol{\gamma}) = \sum_{i = 1}^{N} \alpha\left(\mathbf{x}_{i}\right) \prod_{j<i}\left(1-\alpha\left(\mathbf{x}_{j}\right)\right) \mathbf{c}\left(\mathbf{x}_{i},\mathbf{v}\right),
\label{volume_rendering}
\end{equation}
where $\alpha\left(\mathbf{x}_{i}\right) = 1-\exp \left(-\sigma\left(\mathbf{x}_{i}\right) \delta_{i}\right)$, $\mathbf{x}_{i} = \mathbf{o} + t_{i}\mathbf{v}$ is the uniformly sampled point on the ray, and $\delta_{i} = {t}_{i+1}-{t}_{i}$ are the distance between adjacent sample points. 
To predict the density $\sigma(\mathbf{x})$ and radiance $\mathbf{c}(\mathbf{x}, \mathbf{v})$ at the input point $\mathbf{x}$ from the view direction $\mathbf{v}$, NeRF encodes each query point $\mathbf{x}$ and view direction $\mathbf{v}$ using a positional encoding that projects a coordinate vector into a high-dimensional space. The high-dimensional vectors are then fed into $F_{\omega}$. NeRF optimizes a separate neural network for each scene, and the optimization process usually requires a few hours on a single RTX 3090 GPU, making it time-consuming.


\begin{figure}
  \centering
  \includegraphics[width=\linewidth]{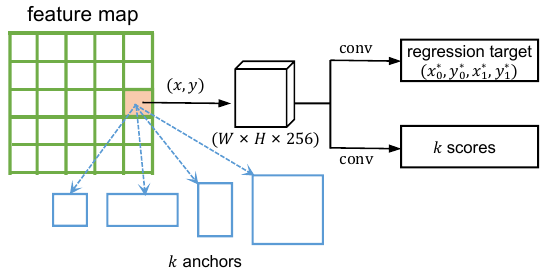}

  \caption{Anchor-free network structure of FCOS.}
  \label{fig:anchor_based}
   \vspace{-0.4cm}
\end{figure}

\noindent{\bf{Multi-resolution Hash Encoding.}}
For improving the training speed of NeRF, Instant-NGP(NGP)\cite{muller2022instant} introduces multi-level hash encoding to replace the positional encoding in NeRF, which improves the convergence speed for reconstructing a 3D scene.
Specifically, NGP maintains $L$-level hash tables, and each table contains $T$ feature vectors with dimensionality $F$. We denote the feature vectors in the hash tables as 
$\mathcal{H} = \left\{{\mathcal{H}}_{l} \mid l \in\{1, \ldots, L\}\right\}$. Each table is independent, and stores feature vectors at the vertices of a grid with resolution $ N_{l}$. 
The number of grids exponentially grows from the coarsest $N_{c}$ to the finest resolution $N_{f}$. Therefore, $N_{l}$ is defined as follows:
\begin{equation}
\begin{array}{c}
b:=\exp \left(\frac{\ln N_{f}-\ln N_{c}}{L-1}\right),\quad \quad \\
N_{l}:=\left\lfloor N_{c} \cdot b^{l}\right\rfloor,
\end{array}
\end{equation}
In practice, we set $N_{c }=16$ and $N_{f}$ is the same as the original resolution of the input images. The growth factor $b^{l}$ is chosen from the interval [1.26,2].

The multi-resolution hash encoding with learnable hash tables $\mathcal{H}$ are denoted as $\mathbf{h}(\cdot|\mathcal{H})$. For given a specific level $l$, an input 3D position $\mathbf{x} \in \mathbb{R}^3$ is scaled by $N_{l}$ and then a grid spans to a unit hypercube by rounding up and down
$ \left\lceil\mathbf{x}_{l}\right\rceil:=\left\lceil\mathbf{x} \cdot N_{l}\right\rceil$, $\left\lfloor\mathbf{x}_{l}\right\rfloor:=\left\lfloor\mathbf{x} \cdot N_{l}\right\rfloor$. 
The feature of $\mathbf{z}$ at level $l$ is tri-linearly interpolated by the feature vectors at the vertices of the diagonal.
As for the coarse resolution, where the number of total vertices is fewer than $T$, the hash table can provide a one-to-one query. However, for higher resolution, the mapping between feature vectors at each vertices and the hash table $\mathcal{H}_{l}$ are established by the following spatial hash function\cite{teschner2003optimized}: 
\begin{equation}
    h(\mathbf{x})=\left(\bigoplus_{i=1}^{3} x_{i} \pi_{i}\right) \bmod T,
\end{equation}
where $\bigoplus$ denotes bitwise XOR operation and $\pi_{i}$ are preset large prime numbers.
The $L$ levels' feature vectors of $\mathbf{x}$ are concatenated to produce $\mathbf{h}(\mathbf{x}|\mathcal{H}) \in \mathbb{R}^{LF}$, which is utilized as the input of MLP.

\begin{figure}
  \centering
  \includegraphics[width=\linewidth]{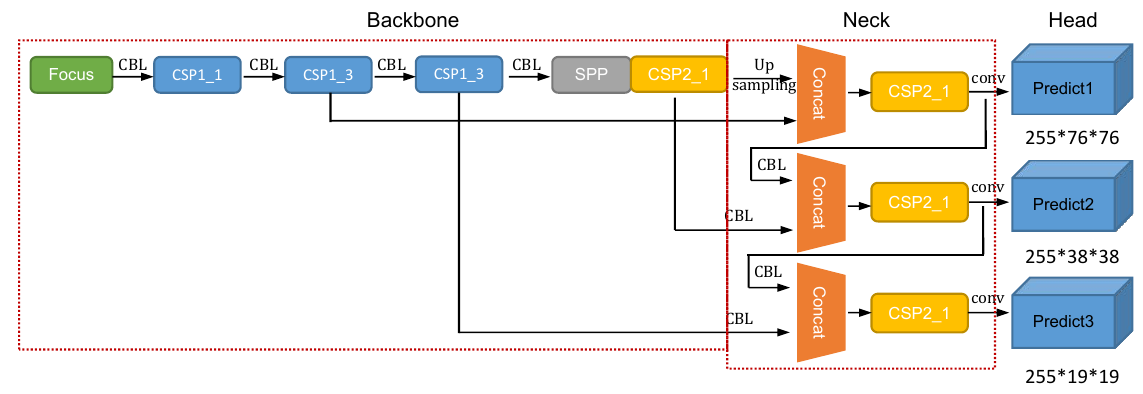}

  \caption{Anchor-based network structure of YOLOv5. 
}
  \label{fig:detection}
   \vspace{-0.4cm}
\end{figure}

\subsection{Incorporating Detectors with Instant Neural Radiance}
\label{sec:Yolov5}
Our Region Proposal Network takes the feature pyramid from the feature extractor and outputs a set of oriented bounding boxes with corresponding objectness scores. We experiment with two types of region proposal methods: anchor-based and anchor-free methods in  Fig.~\ref{fig:detection} and Fig.~\ref{fig:anchor_based}, respectively.

\noindent{\bf{Anchor-Free detector.}} Anchor-free object detectors discard the expensive IoU computation between anchors and ground-truth boxes. We choose the representative anchor-free method - FCOS~\cite{tian2019fcos} which does not require the anchor boxes and proposals. It not only simplifies the computation process associated with anchor boxes but also the hyperparameters associated with anchors.
We define a set of bounding box $B_i = (x^{i}_0, y^{i}_0, x^{i}_1,y^{i}_1, c^i \in (1,2,...,C))$. $(x_0^{i}, y_0^{i})$ and a centerness score $c$. Here $x_1^{i}$ and $ y_1^{i}$ represent the coordinates of the upper-left and lower-right corners of the bounding box. $c^i$ indicates the category that the object in the bounding box belongs to.
The regression target $\boldsymbol B^*_i = (x_0^*, y_0^*, x_1^*, y_1^*)$ can be formulated as following:
\begin{equation}
\begin{alignedat}{2}
    &x_0^* = x - x_0^{(i)},\quad && x_1^* = x_1^{(i)} - x, \\
    &y_0^* = y - y_0^{(i)},\quad && y_1^* = y_1^{(i)} - y, \\
\end{alignedat}
\end{equation}
where $x, y$ are the position on feature map, see Fig.~\ref{fig:detection}. The ground-truth centeredness is given by: 
\begin{equation}
\!\!\!\!    c^* = \sqrt{\frac{\min (x_0^*, x_1^*)}{\max (x_0^*, x_1^*)} \times \frac{\min (y_0^*, y_1^*)}{\max (y_0^*, y_1^*)} }.
\end{equation}
The overall loss is then given by
\begin{align}
\begin{split}
    \!\!\!\!L(\{p_i\}, &\{\boldsymbol B_i\}, \{c_i\}) = \frac{1}{N_{pos}} L_{cls} (p_i, p_i^*) \\ &\!\!\!\!+ \frac{\lambda}{N_{pos}} p_i^* L_{reg} (\boldsymbol B_i, \boldsymbol B_i^*) + \frac{1}{N_{pos}} p_i^* L_{ctr} (c_i, c_i^*),
\end{split}
\label{eq:6}
\end{align}
where $L_{cls}$ is the focal loss in \cite{lin2017focal} and $L_{reg}$ is the IoU loss for rotated boxes in \cite{zhou2019iou}. $L_{ctr}$ is the binary cross entropy loss. $p_i^* \in \{0, 1\}$ is the ground-truth label of each box in the feature pyramid, which is determined using the same center sampling and multi-level prediction process as in \cite{tian2019fcos}; $\lambda$ is the balancing factor and $N_{pos}$ is the number of boxes with $p_i^* = 1$. 


\noindent{\bf{Anchor-based detector.}} YOLOv5 places anchors of different sizes and aspect ratios at each pixel location and predicts objectness scores and bounding box regression offsets for each anchor. Fig.~\ref{fig:detection} demonstrates three main components of the YOLOv5: backbone network, neck network, and head network. It adopts a variant of the CSPNet (Cross-Stage Partial Network) as the backbone network and extracts the feature pyramid. The neck network further refines the features by applying spatial and channel-wise attention, which enhances the feature representation and enables the detection of objects at different scales and aspect ratios. The head network employs a multi-scale prediction scheme, which enables the detection of objects at different scales and resolutions. The head network uses anchor boxes to predict the object's class probability, the offset between the object center and anchor box, and the object size. 
Once the head network generates the output, the predicted bounding boxes are post-processed to eliminate overlapping detections and suppress false positives.

\subsection{Refined Instant-NGP}
\label{sec:hash}
In this part, we detail how to incorporate multi-resolution hash encodings for restoring distracted scenes to accelerate the training process. The primitive Instant-NGP\cite{muller2022instant} is able to rapidly reconstruct 3D scenes but lacks the capability of synthesizing novel camera rays in masked regions. To avoid rendering objects inside bounding boxes, therefore, it is essential to remove the color and density sample points in camera rays that are cast in distractors. 
Suppose $\mathcal{D}_k$ is the set of distractors in image $I_k$, we obtain the corresponding bounding boxes set $\mathcal{M}_k$ by adopting the methods described in Sec.~\ref{sec:Yolov5}. 
Given $\mathbf{r}$ represent the camera rays, we eliminate the concatenated feature vector $\mathbf{h}(\mathbf{r}|\mathcal{H})$ by multiplying a preset weight coefficient in all resolutions, the remaining feature vectors are fed into an MLP $f_\varphi$ to regress radiance and density, which can be defined as:
\begin{equation}
\left(\sigma(\mathbf{r}), \mathbf{c}(\mathbf{r})\right) = f_{\varphi}\left(W_l \times \mathbf{h}(\mathbf{r}|\mathcal{H}_{l})\right), l \in \{1,...,L\},
\end{equation}
\begin{equation}
W_l = 
\begin{cases}
0, & \text{if~} \mathbf{r} \in \mathcal{M}_k \\
1, & \text{otherwise}
\end{cases},
\end{equation}
where $f_{\varphi}$ is the MLP function with learnable parameters ${\varphi}$. $W_k$ is a preset weight coefficient and $\mathcal{H}_{l}$ is the $l$-th level hash tables.
Finally, the color of each sample ray outside the bounding boxes is calculated using the following volume rendering formula:
\begin{equation}
\centering
\mathbf{C}(\mathbf{r}) = \sum_{i = 1}^{N} \alpha(\mathbf{x}^{(i)}_o) \prod_{j<i}(1-\alpha(\mathbf{x}^{(j)}_o)) \mathbf{c}(\mathbf{x}^{(i)}_o),
\label{volume_rendering2}
\end{equation}
where $\alpha(\mathbf{x}^{(i)}_o) = 1-\exp (-\sigma(\mathbf{x}^{(i)}_o) \delta_{i})$. $x^{(i)}_o$ is the sample point of rays $\mathbf{r}$ outside the bounding boxes.

\subsection{Loss Functions}
\label{sec:loss}
We use the following loss function to jointly optimize the network parameters ${\omega, \phi}$ and the feature vectors $\mathcal{H}$ in the hash tables.

\noindent{\bf{Photometric Loss}}.
We minimize the rendering error of all observed images, and the loss function is defined as:
\begin{equation}
\mathcal{L}_{\mathrm{rgb}}=\frac{1}{|\mathcal{R}|}\sum_{\boldsymbol{r} \in \mathcal{R}}\left\|\tilde{\mathbf{C}}(\boldsymbol{r})-\mathbf{C}(\boldsymbol{r})\right\|_{2},
\end{equation}
where $\mathcal{R}$ is the set of rays passing through image pixels outside the bounding boxes and $\tilde{\mathbf{C}}(\boldsymbol{r})$ is the rendered color and ${\mathbf{C}} (\boldsymbol{r})$ is the corresponding ground truth color.

\noindent{\bf{Multi-view Compensation Loss (MVCL)}}.
Since the content of bounding boxes in each view is highly relevant to the corresponding regions in other views, we compensate the weight proportion of the corresponding regions to reinforce the smoothness of reconstruction results.
we sample another ray $\boldsymbol{r}$ in the same position of different viewpoint. Then, the multi-view compensation loss is
\begin{equation}
\mathcal{L}_{\mathrm{MVCL}}=\frac{s \times n_r}{|\mathcal{R}|}\sum_{\boldsymbol{r} \in \mathcal{R}}\left\|\tilde{\mathbf{C}}(\boldsymbol{r})-\mathbf{C}(\boldsymbol{r})\right\|_{2},
\end{equation}
where $n_r$ is the number of regions without covering by bounding boxes corresponding to observed ray $\boldsymbol{r}$ and $s$ is the scale factor. $n_r$ is denoted as follows:
\begin{equation}
n_r= \sum_{M_k \in \mathcal{M}}\mathbb{I}(\boldsymbol{r} \notin \boldsymbol{M}_k),
\end{equation}
where $M_k$ is the region inside bounding boxes of the $k$-th views and $\mathcal{M}$ is the total corresponding regions inside bounding boxes.

\noindent{\bf{Perceptual Loss}}. We adopt learned perceptual image patch similarity (LPIPS) to build the perceptual loss $L_{\mathrm{LPIPS}}$, which is utilized to provide robustness to slight distortions and shading variations and improve details in the reconstruction. We choose the pre-trained VGG network~\cite{johnson2016perceptual} as the backbone of LPIPS to extract features in each patch. The rendered patch is compared with the same position patch on the input image, which is defined as: 
\begin{equation}
\mathcal{L}_{\mathrm{LPIPS}}=\left\|{F_{\text{vgg}}}(\tilde{\boldsymbol{I}})-F_{\text{vgg}}(\boldsymbol{I})\right\|_{2},
\end{equation}
where $\tilde{\boldsymbol{I}}$ is the rendered image patch, $\boldsymbol{I}$ is the ground truth image.
The total loss function is formulated as:
\begin{equation}
     \mathcal{L}_{\mathrm{total}} = \mathcal{L}_{\mathrm{rgb}} +\lambda_{1}\mathcal{L}_{\mathrm{MVCL}} + \lambda_{2}\mathcal{L}_{\mathrm{LPIPS}},
\end{equation}

We set $\lambda_1=0.01$ and $\lambda_2=0.1$ for the first 400 iterations to learn a coarse result and then increase $\lambda_1$ and $\lambda_2$ to learn the details in bounding boxes.


\begin{figure*}
	\centering
	\scalebox{1}{\includegraphics[width=1.0\linewidth]{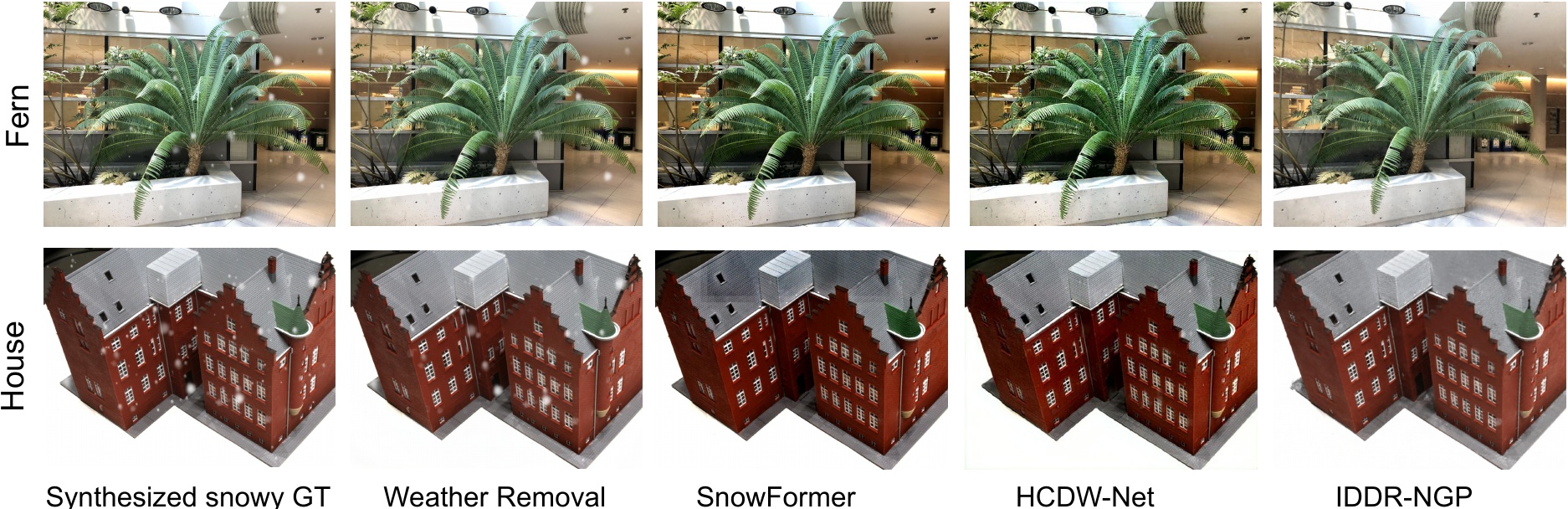}}
	
	\caption{
		 Qualitative results on synthetic scenes. Given the input images, we visualize the desnow results of Weather removal, SnowFormer, HCDW-Net and our IDDR-NGP in a consistent view. 
	}
	\label{fig:desnow}
\end{figure*}

\begin{table}
\caption{Quantitative results on DTU and LLFF datasets. We add synthetic snowflakes to each scene and evaluate the average score. The best results are presented in bold. 
}
\centering
\footnotesize
\setlength{\tabcolsep}{1pt}
\newcolumntype{Y}{>{\centering\arraybackslash}X}
\begin{tabularx}{0.999\linewidth}{l||YYY|YYY|YYY}
\toprule

 & \multicolumn{3}{c}{\scshape  Fern}
 & \multicolumn{3}{c}{\scshape  House}

\\
  \textsl{Method} & \multicolumn{1}{c}{\scriptsize PSNR$\uparrow$}    & \multicolumn{1}{c}{\scriptsize SSIM$\uparrow$} & \multicolumn{1}{c}{\scriptsize LPIPS$\downarrow$} & \multicolumn{1}{c}{\scriptsize PSNR$\uparrow$} & \multicolumn{1}{c}{\scriptsize SSIM$\uparrow$} & \multicolumn{1}{c}{\scriptsize LPIPS$\downarrow$} \\
  \midrule
\scriptsize \textit{Weather removal}                  &24.22      &0.71      &0.43   &23.73        &0.69    &0.49          \\
 \scriptsize \textit{SnowFormer}                    &28.34      & 0.87    & 0.31   & 27.93        & 0.75   &  0.41           \\
 \scriptsize \textit{HCDW-Net}                   & 28.17    & \textbf{0.94}    & 0.24     & 27.31   & \textbf{0.92}    & 0.26            \\

\scriptsize \textit{IDDR-NGP}                 & \textbf{31.74}    & 0.92    & \textbf{0.25}    & \textbf{31.27}        & 0.91    & \textbf{0.30}           \\
\midrule
\bottomrule
\end{tabularx}

\label{tab:Desnow quantitative results}
\vspace{-0.2in}
\end{table}

\section{Performance Evalutaion}
In this section, we perform comparative experiments on both synthetic and realistic scenes with simulated snowflakes to demonstrate the effectiveness of IDDR-NGP. Furthermore, we conduct experiments on different distractors to evaluate the robustness and generalization of our framework. Some ablation studies are discussed to evaluate the compatibility of our detection modules and the effectiveness of proposed loss function. 
\subsection{Datasets and Metrics}
\noindent{\bf{Datasets.}} Since there are no corrupted multi-view 3D scene datasets for evaluating the restoration capability of our method, we obtain our dataset by capturing and synthesizing distractors on both realistic and synthetic scenes. 
The synthetic scenes include DTU and LLFF datasets with snow, petals, and defoliation as the added distractor. 
Our realistic scenes are captured by an Apple iPhone 12, which contains 20-30 images with and without confetti as distractors, and each image has a resolution of $1276 \times 1276$.
\noindent{\bf{Metrics.}} We use standard metrics to quantitatively evaluate our synthesis results: peak signal-to-noise ratio (PSNR) and structural similarity index (SSIM). To reduce the effect of the background, we compute these metrics only for the pixels inside the 2D bounding box obtained from detectors for each view.

\subsection{Implementation Details}
\label{detail}
We implement our method with the torch-ngp\footnote{https://github.com/ashawkey/torch-ngp} codebase~\cite{tiny-cuda-nn,muller2022instant}.
We optimize with Adam~\cite{kingma2014adam} using a learning rate decay from $2 \cdot 10^{-3}$ to $2 \cdot 10^{-5}$. Following Instant-NGP~\cite{muller2022instant}, the multi-resolution hash encoder has 16 levels, each returning two features. The base resolution is set to 16. The runtimes of COLMAP and Instant-NGP are around 15 minutes and 45 minutes with 90000 iterations. 
The convergence time of our method depends on the length and resolution of the input images.
For $1080 \times 1080$ resolution images, we need about 30000 iterations to converge (about 15 minutes on a single NVIDIA GeForce RTX3090).
In addition, our model requires extra memory to store the hash tables because we use multi-resolution hash encoding to speed up training. 
However, it does not consume an excessive amount of graphics memory.
The hyper-parameters and implementation of detectors are same with FCOS~\cite{tian2019fcos} and YOLOv5~\cite{zhu2021tph}.
Specifically, the backbone network is trained with SGD for 300 epochs with the initial learning rate being 0.01 and a mini-batch of 16 images. Weight decay and momentum are set as 0.0001 and 0.9, respectively.

\begin{figure*}
	\centering
	\scalebox{1}{\includegraphics[width=1.0\linewidth]{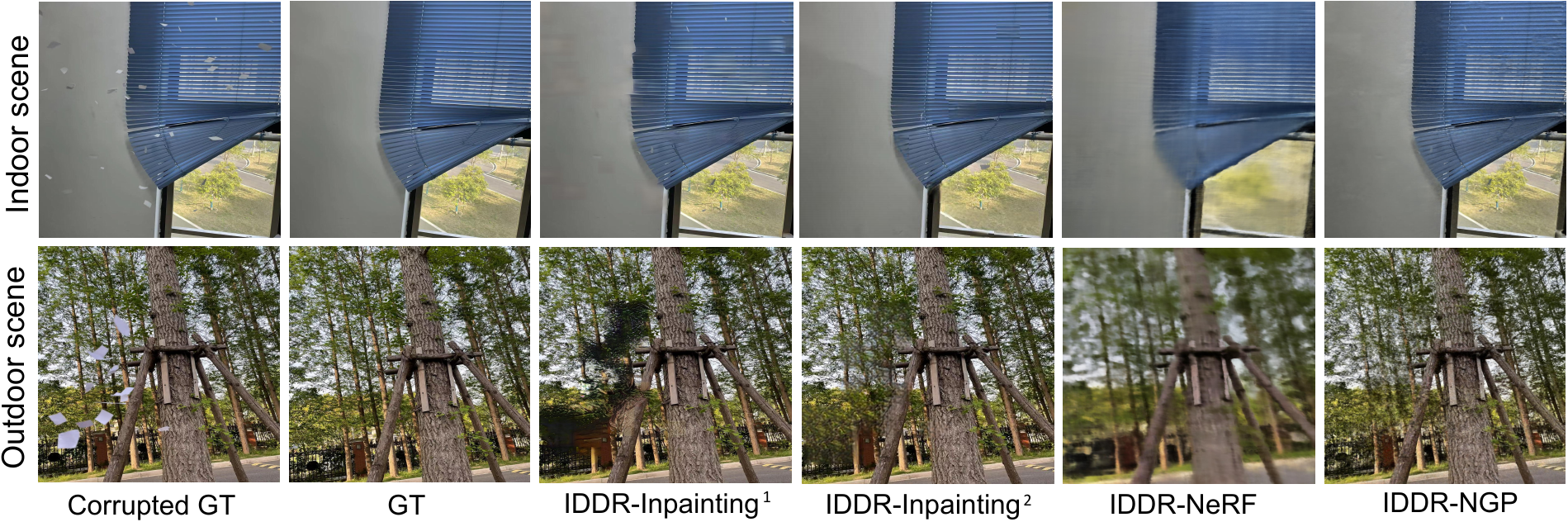}}
	
	\caption{
		 Qualitative results on realistic scenes with real distractors. We compare our method with IDDR-Inpainting$^1$ and IDDR-Inpainting$^2$. Note that CR-Fill is utilized as the inpainting backbone in IDDR-Inpainting$^1$ and T-Fill is adopted as the inpainting backbone in IDDR-Inpainting$^2$.
	}
	\label{fig:realistic_scene}
 \vspace{-0.4cm}
\end{figure*}

\subsection{Results on Synthetic Scene with Synthetic Distractors}
To demonstrate the superior performance of our manner on synthetic datasets, we select two scenes from the DTU, and LLFF datasets, respectively. Then, we test our IDDR-NGP by adding synthetic snow to these benchmarks.

\noindent{\bf{Qualitative Results.}}
We compare the visual performance of IDDR-NGP with state-of-the-art methods for snow removal tasks on synthesized snowy scenes, which are presented in Fig.~\ref{fig:desnow}. Weather removal~\cite{chen2022learning} is a unified model for multiple adverse weather and struggles to remove the snowflakes in real scenes.
Although the latest desnowing methods are able to completely remove all snow in target scenes due to ignoring multi-view context information, especially large snow pieces. Similarly, the overall textural color are change in SnowFormer~\cite{chen2022snowformer} are not recovered well. As observed in Fig.~\ref{fig:desnow}, it is visually found that the recovered results by SnowFormer still have residual snow spots and snow marks with small sizes. 
Although HCDW-Net~\cite{chen2021all} can remove most snow particles, snowflakes with small sizes cannot be removed effectively. 
In contrast, compared with these SOTA methods, IDDR-NGP provides more pleasant snow removal results with more details, indicating superior generalization ability in various scenes, and the restored images are closer to the ground truths. 



\noindent{\bf{Quantitative Results.}} 
For a fair comparison, we retrain each compared De-snow model based on our synthesized snowy multi-view dataset and report the best result in Tab.~\ref{tab:Desnow quantitative results}. 
Since the traditional snow removal algorithm processes a single image, we calculate the average scores of PSNR, SSIM, and LPIPS in RGB color space for multi-view images to evaluate the snow removal performance for all experiments.

Tab.~\ref{tab:Desnow quantitative results} summarizes the quantitative comparisons between Desnow-Net, SnowFormer, HDCW-Net, and IDDR-NGP, which indicates that IDDR-NGP outperforms other single-view independent de-snow methods on all the presented datasets.

\begin{table}
\caption{Quantitative results on realistic scenes with confetti as distractors. We calculate the average score over all multi-view images. The best results are presented in bold.
}
\label{tab:realistic_scene}
\centering
\footnotesize
\setlength{\tabcolsep}{1pt}
\newcolumntype{Y}{>{\centering\arraybackslash}X}
\begin{tabularx}{0.999\linewidth}{l||YYY|YYY}
\toprule
& \multicolumn{3}{c}{\scshape Indoor scene }
 & \multicolumn{3}{c}{\scshape Outdoor scene  }

\\
  & \multicolumn{1}{c}{\scriptsize PSNR$\uparrow$}    & \multicolumn{1}{c}{\scriptsize SSIM$\uparrow$} & \multicolumn{1}{c}{\scriptsize LPIPS$\downarrow$} & \multicolumn{1}{c}{\scriptsize PSNR$\uparrow$} & \multicolumn{1}{c}{\scriptsize SSIM$\uparrow$} & \multicolumn{1}{c}{\scriptsize LPIPS$\downarrow$}  \\
  \midrule
\scriptsize \textit{IDDR-Inpainting$^1$}              & 16.26 & 0.83 & 0.42     &  15.32    & 0.84     & 0.37          \\
\scriptsize \textit{IDDR-Inpainting$^2$}              & 20.91 & 0.86 & 0.33     &  21.66    &0.87     & 0.33           \\
\scriptsize \textit{IDDR-NeRF}              & 18.32 & 0.77 & 0.53     &  17.74    &0.63     & 0.58           \\

\scriptsize \textit{IDDR-NGP}         & \textbf{28.41} & \textbf{0.84} & \textbf{0.31}   &  \textbf{33.27}    & \textbf{0.92}    & \textbf{0.21}   \\

\midrule
\bottomrule
\end{tabularx}

\vspace{-0.4cm}
\end{table}

\subsection{Results on Realistic Scene with Realistic Distractors}
To further evaluate the effectiveness of our method on real-world scenes with real distractors, we choose two different types of scenes for realistic scene distractors removal, including indoor and outdoor scenes, and conduct experiments by scattering confetti to the target scenes. Furthermore, a defined baseline, denoted as \textbf{IDDR-Inpainting}, is proposed to compare with our \textbf{IDDR-NGP}. Specifically, we directly infill the region inside the bounding boxes with CR-Fill~\cite{zeng2021cr} and TFill~\cite{zheng2022bridging}, respectively.

\noindent{\bf{Qualitative Results.}} 
Qualitative results are shown in Fig.\ref{fig:realistic_scene}, which indicates that the results of IDDR-Inpainting demonstrate blurry outputs and inconsistent inpainted results in regions inside bounding boxes. In contrast, compared with self-defined IDDR-Inpainting, our approach produces photo-realistic renderings and visual-consistent results. In addition, our results are almost invisible for the confetti that initially floated on the curtains in the \emph{Indoor} scene. The same observation was performed on the \emph{Outdoor} scene.

\noindent{\bf{Quantitative Results.}} 
Quantitative results are shown in Tab.~\ref{tab:realistic_scene}, our method significantly outperforms IDDR-Inpainting$^2$ in terms of PSNR and LPIPS. IDDR-NGP achieves respectively 36\%, and 54\% superior to IDDR-Inpainting$^2$ in PSNR. Similar improvements can be seen in SSIM and LPIPS metrics. Although the performance of IDDR-Inpainting is close to that of IDDR-NGP, its output lacks any geometry guidance and multi-view context information which demonstrate difficulty in generating the view-consistent results.

\begin{figure*}
	\centering
	\scalebox{1}{\includegraphics[width=1.0\linewidth]{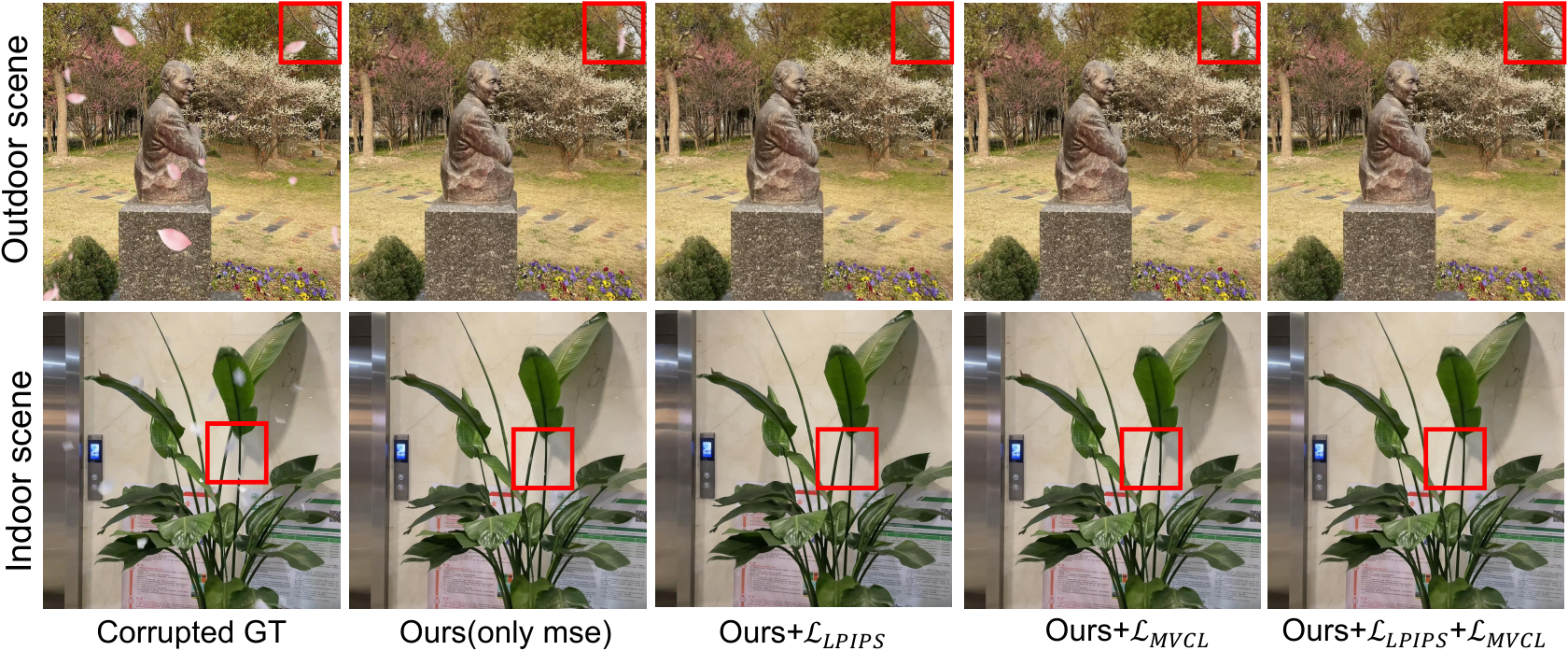}}
	
	\caption{
		 Ablation results of our proposed loss function on synthetic and realistic scenes. From left to right, we begin the experiment with only mse. Then we add $\mathcal{L}_{LPIPS}$ and $\mathcal{L}_{MVCL}$ as order. Both $\mathcal{L}_{LPIPS}$ and $\mathcal{L}_{MVCL}$ improves the final results. 
	}
	 \label{fig:Ablation_loss}
\end{figure*}

\subsection{Ablation Study}
\noindent{\bf{Ablation Study on the Multi-resolution Hash Encoding.}}
In this part, we design the following baseline \textbf{IDDR-NeRF} to further compare the effects of hash encoding in our method. Specifically, we train the primitive NeRF model with the same pipeline and equal rays in the target scene. The corresponding rendering results of IDDR-NeRF are shown in Fig.~\ref{fig:realistic_scene}. 
Notably, we train IDDR-NGP and IDDR-NeRF with same number of rays, however, our method takes fewer train hours to restore scenes; while the training time of IDDR-NeRF for each scene is about 10 hours, which is demonstrated on Tab.~\ref{tab:realistic_scene}.
This representation takes at least twenty times longer to converge. Using multi-resolution hash encoding dramatically increases the training speed of our method.

\begin{table}
\caption{Comparision with NeRF as Backbone on \emph{confetti} scene. The corresponding rendering results are $\mathcal{L}_{LPIPS}$ and w/o $\mathcal{L}_{MVCL}$, respectively.}
\label{tab:Ablation_loss}
\centering
\footnotesize
\setlength{\tabcolsep}{1pt}
\newcolumntype{Y}{>{\centering\arraybackslash}X}
\begin{tabularx}{0.999\linewidth}{l||YYY|YYY}
\toprule
& \multicolumn{3}{c}{\scshape Outdoor Scenes }
 & \multicolumn{3}{c}{\scshape Indoor Scenes }

\\
  & \multicolumn{1}{c}{\scriptsize PSNR$\uparrow$}    & \multicolumn{1}{c}{\scriptsize SSIM$\uparrow$} & \multicolumn{1}{c}{\scriptsize LPIPS$\downarrow$} & \multicolumn{1}{c}{\scriptsize PSNR$\uparrow$} & \multicolumn{1}{c}{\scriptsize SSIM$\uparrow$} & \multicolumn{1}{c}{\scriptsize LPIPS$\downarrow$}  \\
  \midrule
A:\scriptsize \textit{IDDR-NGP}    & 30.66 & 0.94 & 0.26    & 28.17    &  0.92    &  0.32          \\
B: A + $\mathcal{L}_{LPIPS}$     & 30.76 & 0.95 & 0.25    & 27.97    & 0.91     &  0.28          \\
C: A + $\mathcal{L}_{MVCL}$     & 31.04 & 0.95 & 0.25    & 29.52    &  0.92    &  0.33          \\
D: B + $\mathcal{L}_{MVCL}$         & \textbf{32.58} & \textbf{0.96} & \textbf{0.24}   &  \textbf{30.18}    & \textbf{0.93}    & \textbf{0.27}   \\

\midrule
\bottomrule
\end{tabularx}
\vspace{-0.8cm}
\end{table}

\noindent{\bf{Effectiveness of Loss Function.}} 
Here we further evaluate the effectiveness of the proposed loss function in IDDR-NGP on \emph{Indoor} scene. As discussed in Sec.~\ref{sec:loss}, the goal of $\mathcal{L}\text{total}$ is to leverage global and multi-view information to optimize the region inside distractors from multiple perspectives.
Tab.~\ref{tab:Ablation_loss} shows that with the addition of $\mathcal{L}_{MVCL}$, the PSNR metrics are significantly increased due to the multi-view information referenced. Furthermore, the added $\mathcal{L}_{LPIPS}$ is able to exploit the feature information to recover the entire scene. So the best result is achieved when $\mathcal{L}_{LPIPS}$ and $\mathcal{L}_{MVCL}$ are jointly adopted. Fig.~\ref{fig:Ablation_loss} illustrates that our method eliminates undesirable distractors with the collaboration of these two losses and retains 3D consistency in the 2D bounding boxes.

\noindent{\bf{Ablation Study with Different Types of Distractors.}} 
We conduct experiments to validate the robustness and generalization of IDDR-NGP with different types of distractors on both synthetic and realistic scenes. Qualitative results are present in Fig.~\ref{fig:teaser}, which indicates that our method works well when the distractors are various changes. Note that the black background on Blender is transparent, therefore the result of \emph{Ship} performs blurry due to the missing of large information in 360° views.

\section{Discussions}

\noindent{\bf{Failure cases.}}
Our proposed IDDR-NGP can not be utilized for large-scale static distractors removal. 
For example, our method can not remove the distractors that are stationary in multiple views. 

\noindent{\bf{Limitations and Future work.}}
Currently, our method relies on the accuracy of detection methods, and it isn't easy to obtain accurate enough mask labels from monocular videos. This problem might be solved by the recently popular method - the Segment Anything Model~(SAM)~\cite{kirillov2023segment}, which we leave as future work. 
In realistic scenarios, our approach is limited by the performance of target detection methods. We will explore more advanced detection algorithms to enable IDDR-NGP to perform superior results in real-world scenarios. On the other hand, exploring ways to detach from reliance on detection algorithms is a fascinating approach to automatic distractors removal.

\vspace{-0.4cm}

\section{Conclusion}
In this paper, we propose IDDR-NGP, an efficient distractor removal method for 3D scenes by incorporating implicit 3D representation with 2D detectors. 
We modify the rendering strategy in primitive Instant-NGP and introduce the LPIPS and MVCL loss functions to jointly optimize the rendering results of IDDR-NGP, which can aggregate information from multi-view corrupted images. To verify the effectiveness of IDDR-NGP, we provide a wide range of distractors with corresponding annotated labels added to both realistic and synthetic scenes.
Extensive experiments validate the superiority of IDDR-NGP in removing multiple types of distractors. In addition, our method achieves comparable results with existing SOTA desnow methods and is capable of precisely removing both realistic and synthetic distractors.

\noindent \textbf{Acknowledgement}. Jihong Guan was supported by National Natural Science Foundation of China (NSFC) under grant No.~U1936205.
Shuigeng Zhou was partially supported by Open Research Projects of Zhejiang Lab (No. 2021KH0AB04).

\bibliographystyle{ACM-Reference-Format}
\balance
\bibliography{reference}


\end{document}